\documentclass[letterpaper]{article} 
\usepackage{aaai23}  
\usepackage{times}  
\usepackage{helvet}  
\usepackage{courier}  
\usepackage[hyphens]{url}  
\usepackage{graphicx} 
\usepackage{natbib}  
\usepackage{caption} 
\usepackage{algorithm}
\usepackage{algorithmic}
\usepackage{tabularx}
\usepackage{booktabs}
\usepackage{array}
\usepackage{amsmath}
\usepackage{subfigure}
\usepackage{multirow}
\usepackage{bm}
\usepackage{cleveref}
\usepackage{amssymb}
\usepackage{newfloat}
\usepackage{listings}
\DeclareCaptionStyle{ruled}{labelfont=normalfont,labelsep=colon,strut=off} 
\floatstyle{ruled}
\newfloat{listing}{tb}{lst}{}
\floatname{listing}{Listing}
\title{Graph Anomaly Detection via Multi-Scale Contrastive \\ Learning Networks with Augmented View}

\author{
    Jingcan Duan,\textsuperscript{\rm 1} Siwei Wang,\textsuperscript{\rm 1} Pei Zhang,\textsuperscript{\rm 1} En Zhu,\textsuperscript{\rm 1}\thanks{Corresponding author}\\
    Jingtao Hu,\textsuperscript{\rm 1} Hu Jin,\textsuperscript{\rm 1} Yue Liu,\textsuperscript{\rm 1} Zhibin Dong\textsuperscript{\rm 1}
}
\affiliations{
    \textsuperscript{\rm 1}College of Computer, National University of Defense Technology, Changsha, China\\
    \{jingcan\_duan, yueliu19990731\}@163.com,  \{wangsiwei13, zhangpei, enzhu, hujingtao17, jinhu, dzb20\}@nudt.edu.cn

}

\usepackage{bibentry}

\begin{document}

\maketitle

\begin{abstract}
Graph anomaly detection (GAD) is a vital task in graph-based machine learning and has been widely applied in many real-world applications. The primary goal of GAD is to capture anomalous nodes from graph datasets, which evidently deviate from the majority of nodes. Recent methods have paid attention to various scales of contrastive strategies for GAD, i.e., node-subgraph and node-node contrasts. However, they neglect the subgraph-subgraph comparison information which the normal and abnormal subgraph pairs behave differently in terms of embeddings and structures in GAD, resulting in sub-optimal task performance. In this paper, we fulfill the above idea in the proposed multi-view multi-scale contrastive learning framework with subgraph-subgraph contrast for the first practice. To be specific, we regard the original input graph as the first view and generate the second view by graph augmentation with edge modifications. With the guidance of maximizing the similarity of the subgraph pairs, the proposed subgraph-subgraph contrast contributes to more robust subgraph embeddings despite of the structure variation. Moreover, the introduced subgraph-subgraph contrast cooperates well with the widely-adopted node-subgraph and node-node contrastive counterparts for mutual GAD performance promotions. Besides, we also conduct sufficient experiments to investigate the impact of different graph augmentation approaches on detection performance. The comprehensive experimental results well demonstrate the superiority of our method compared with the state-of-the-art approaches and the effectiveness of the multi-view subgraph pair contrastive strategy for the GAD task.
\end{abstract}

\begin{figure*}
    \centering
    \includegraphics[width =0.90 \textwidth]{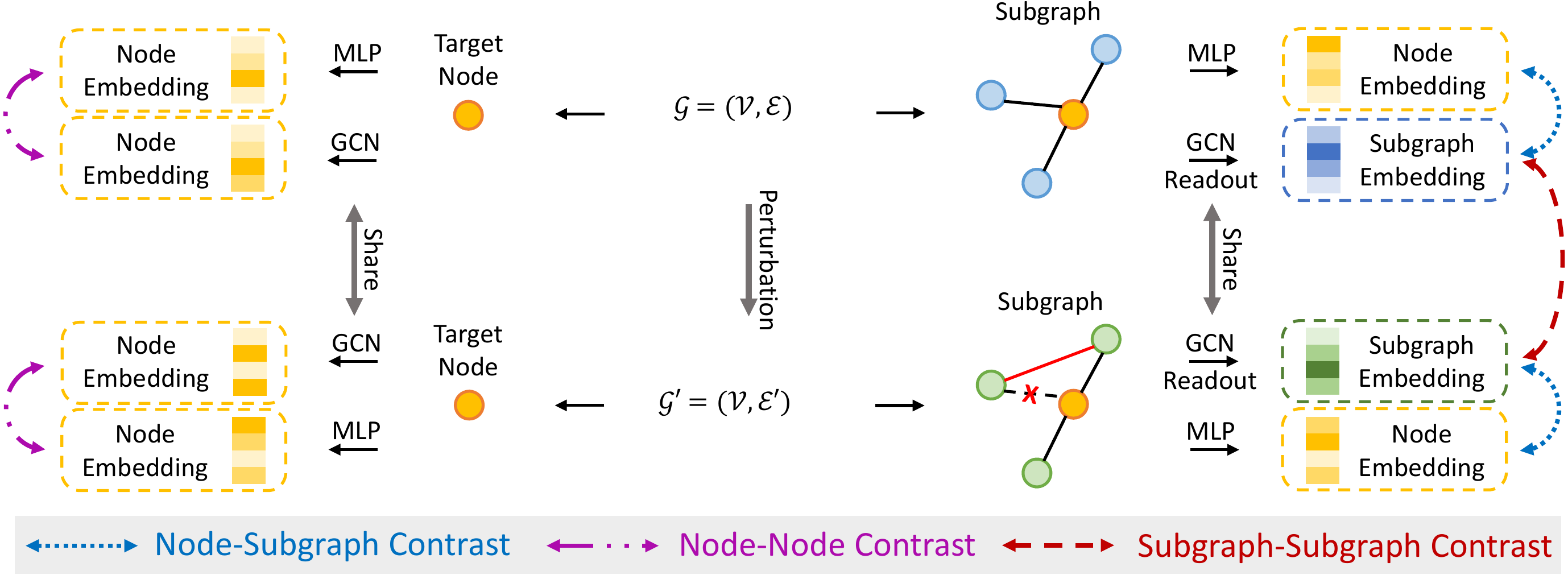}
    \caption{Overview of the GRADATE model. It is composed of two main modules: (1) Graph augmentation. We take the original graph as the first view and the edge-modified graph as the second view. Subgraphs used in the later module are generated by random walk with restart. (2) Graph contrastive network. The network captures various anomalous information from multiple scales under two views by building node-subgraph, node-node, and subgraph-subgraph contrasts. Then we comprehensively calculate the anomaly score for each node.}
    \label{fig:model}
\end{figure*}

\section{Introduction}

Over the past few years, graph-based machine learning has attracted great attention~\cite{wu2020comprehensive, yue2022survey}. As a representative task in graph learning, graph anomaly detection, which intends to discern the anomalies from the majority of nodes, is becoming an increasingly glamorous application for researchers~\cite{ma2021comprehensive}. Due to its vital value in the prevention of harmful events, GAD has already been widely used in many fields, e.g., misinformation detection~\cite{wu2019misinformation}, financial fraud detection~\cite{huang2018codetect}, network intrusion detection~\cite{garcia2009anomaly}, etc. Unlike the data from other anomaly detection fields~\cite{cheng2021improved, cheng2021unsupervised, hu2022detecting}, graph data includes node features and graph structure. The mismatch between these two types of information produces two typical anomalous nodes, i. e. feature anomalies and structure anomalies~\cite{liu2021anomaly}. The former refers to nodes that differ from their neighbors in terms of features, while the latter refers to a group of nodes that are dissimilar but closely connected.

To detect these two categories of anomalies, many previous methods have made great efforts and achieved impressive results. LOF~\cite{breunig2000lof} obtains anomalous information about a node by comparing it with its contextual nodes in terms of features. SCAN~\cite{xu2007scan} accomplishes the GAD task from network structure. By utilizing such two types of information, ANOMALOUS~\cite{peng2018anomalous} detects anomalies based on CUR decomposition and residual analysis. \cite{muller2013ranking, perozzi2014focused} perform feature subspace selection and find anomalous nodes in the subspace. The methods above rely on specific domain knowledge and cannot mine deep non-linear information in graph datasets. These make it difficult for them to improve the detection performance further.

Thanks to the powerful graph information acquisition capability, graph convolutional networks (GCN)~\cite{kipf2016semi} have recently achieved excellent performance in many graph-data tasks. It is naturally applied to detect anomalies in graph. The pioneering work DOMINANT~\cite{ding2019deep} introduces GCN to accomplish the task for the first practice. Specifically, DOMINANT compares the restructured feature and adjacency matrices with the original input matrices. The nodes with more considerable variations have a higher probability of being anomalies. Although it performs well and is simple to implement, some anomalous information will be ignored. GCN generates node representations by aggregating the information from neighborhoods, which will make anomalies more indistinguishable~\cite{tang2022rethinking}. Based on the contrastive learning paradigm, CoLA~\cite{liu2021anomaly} detects anomalies by calculating the relationship between nodes and their neighborhoods. This approach digs the local feature and structure information around the nodes. Meanwhile, it masks the features of the target node during training, which alleviates representation averaging. Differently, ANEMONE~\cite{jin2021anemone} adds the node-node contrast to contrastive networks and focuses on the node-level anomalous information.

However, existing works neglect further exploitation of subgraph information and do not directly optimize their embeddings for graph anomaly detection. \cite{jiao2020sub, hafidi2022negative, han2022generative} have demonstrated that subgraph representation learning benefits graph-based machine learning tasks. It will significantly facilitate mining local features and structure information for respective subgraphs. For GAD, more representative and intrinsic subgraph embeddings can help to compute a more reliable relationship between nodes and their neighborhoods, which is the crucial step in the contrastive strategy.

To tackle the issue, we propose a new \textbf{GR}aph \textbf{A}nomaly \textbf{D}etection framework via multi-scale contrastive learning networks with newly added subgraph-subgraph contrast and \textbf{A}ugmen\textbf{TE}d view (termed \textbf{GRADATE}). To be specific, we regard the original input graph as the first view and adopt edge modification as graph augmentation technology to generate the second view. In each view, subgraphs are sampled by random walk. Then we build a multi-view contrastive network with node-subgraph, node-node, and subgraph-subgraph contrasts. The first two contrasts can capture subgraph-level and node-level anomalous information from each view. Subgraph-subgraph contrast is defined between two views and digs more local anomalous information for detection. In this way, node-subgraph contrast will be evidently enhanced. After that, we combine various anomalous information and calculate the anomaly score for each node. Finally, we explore and analyze the effect of different graph augmentation on subgraph representation learning in GAD. Our main contributions are listed as follows:

\begin{itemize}
    \item We introduce subgraph-subgraph contrast to GAD for the first practice and propose a multi-scale contrastive learning networks framework with an augmented view.
    \item We investigate the effects of different graph augmentation on subgraph representation learning for the task.
    \item Extensive experiments on six benchmark datasets prove the effectiveness of the edge modification-based subgraph-subgraph contrast strategy for graph anomaly detection and the superiority of GRADATE comparing to the state-of-the-art methods.
\end{itemize}

\section{Related Work}
\subsection{Graph Anomaly Detection}
The early works~\cite{li2017radar, perozzi2016scalable, peng2018anomalous} usually adopt the non-deep paradigm, which detect the anomalous information from either node features and network structure. However, they cannot continuously improve their performance without digging deeper information. In recent years, the rise of neural networks~\cite{DFCN2021, ITR2022, KGESymCL, yue2022survey} has enhanced the capability of models to mine deep nonlinear information. The reconstruction-based approach DOMINANT~\cite{ding2019deep} obtains node anomaly scores by calculating variations of the feature and structure matrices after GCN. AAGNN~\cite{zhou2021subtractive} applies the one-class SVM in graph anomaly detection. HCM~\cite{huang2021hop} regards the hop-count estimate of the node and its one-order neighbors as its anomaly score. CoLA~\cite{liu2021anomaly} firstly introduces the contrastive learning paradigm~\cite{yang2022interpolation, 9817089} to detect node anomalies in graph. Later methods~\cite{jin2021anemone, zheng2021generative, zhang2022reconstruction, jingcan2022gadmsl} make further improvements based on CoLA.

\subsection{Graph Constrastive Learning}
Contrastive learning is one of the most crucial paradigms in unsupervised learning. Graph contrastive learning excavates supervised information for downstream tasks without expensive labels and has made great achievements~\cite{liu2022graph}. Based on the negative sample usage strategy, the existing works can be divided into negative-based and negative-free Subcategories. For the first type, DGI~\cite{velickovic2019deep} maximizes the mutual information between nodes and graphs to obtain useful supervised information. SUBG-CON~\cite{jiao2020sub} and GraphCL~\cite{hafidi2022negative} form node-subgraph contrasts to learn better node representations. For the second type, BGRL~\cite{thakoor2021bootstrapped} applies a siamese network to draw affluent information from two views. \cite{liu2022deep, liu2022simple, liu2022improved} takes the advantage of Barlow Twins~\cite{zbontar2021barlow}, which designs a special loss function to avoid representation collapse.

\subsection{Graph Augmentation}
Graph augmentation produces plausible variations of graph datasets. It expands the datasets and improves the generalization capability of the model without expensive labels~\cite{zhao2022graph}. Most methods focus on the operations for nodes or edges in graph. \cite{wang2021mixup, feng2020graph, you2020graph} pay attention to modify the node features. RoSA~\cite{zhu2022rosa} uses random walk with restart as the graph augmentation to learn the robust representation of nodes. \cite{klicpera2019diffusion, zhao2021data} adjust the adjacency matrix by adding or removing edges.

\begin{table}[t]
\centering
\caption{Notation summary}
\begin{tabularx}{0.45\textwidth}{p{2.2cm}<{\centering}p{5.3cm}}
\toprule
$\textbf{Notations}$ & $\textbf{Definitions}$\\
\midrule
$\mathcal{G}$       & An undirected attributed graph\\
$v_{i}$       & The $i$-th node of $\mathcal{G}$\\
$\mathbf{A}\in\mathbb{R}^{n\times n}$       & The adjacency matrix of $\mathcal{G}$\\
$\mathbf{D}\in\mathbb{R}^{n\times n}$&The degree matrix of $\mathbf{A}$\\
$\mathbf{X}\in\mathbb{R}^{n\times d}$       & The feature matrix of $\mathcal{G}$\\
$\mathbf{H}^{\left ( \ell \right ) }\in\mathbb{R}^{n\times d^{\prime}}$&The ${\ell}$-th layer hidden representation matrix\\
$\boldsymbol{h}_{i}^{\left ( \ell \right ) }\in\mathbb{R}^{1\times d^{\prime}}$     & The ${\ell}$-th layer hidden representation of $v_{i}$\\
$\mathbf{W}^{\left ( \ell \right )}\in\mathbb{R}^{d^{\prime}\times d^{\prime}}$&The ${\ell}$-th layer network parameters\\

$R$&The number of anomaly detection rounds\\
$S_{i}$&The final anomaly score of $v_{i}$\\
\bottomrule
\end{tabularx}
\label{table:symbol}
\end{table}

\section{Problem Definition}
In the following section, we formalize the graph anomaly detection task. For the given undirected graph $\mathcal{G} = \left ( \mathcal{V}, \mathcal{E} \right ) $, $\mathcal{V} = \left \{ v_{1}, v_{2},..., v_{N} \right \} $ and $\mathcal{E}$ indicate the node set with $N$ nodes and the edge set with $M$ edges, separately. In addition, the node feature matrix $\mathbf{X} \in \mathbb{R}^{n\times d}$ and the adjacency matrix $\mathbf{A} \in \mathbb{R}^{n\times n}$ contain the node feature information and the graph structure information. Here, $\mathbf{A}_{ij} = 1$ means there is an edge between node $v_{i}$ and $v_{j}$, otherwise $\mathbf{A}_{ij} = 0$. The graph anomaly detection model works for learning an anomaly scoring function $f$, which estimates the anomaly score $S_{i}$ for each node. The larger $S_{i}$ represents that the node is more likely to be anomalous. Table~\ref{table:symbol} demonstrates the main used notations in this paper.

\section{Method}
In this section, we introduce the proposed framework, GRADATE. It consists of two main modules. In the graph augmentation module, we treat the original graph as the first view and produce the second view by edge modification. In the graph contrastive network module, we first obtain anomalous information from the feature comparison of nodes and subgraphs, regarded as node-subgraph contrast. For each view, subgraphs are sampled by random walk and form pairs with the target node. Then we build node-node contrast to capture node-level anomalies. The newly added subgraph-subgraph contrast directly optimizes the subgraph embeddings for GAD between two views. In this process, the performance of node-subgraph contrast will be boosted. After that, we adopt an integrated loss function to train these three contrasts. Finally, we synthesize various anomalous information and compute the anomaly score of each node.

\subsection{Graph Augmentation}
Graph augmentation is crucial for the self-supervised learning paradigm. It can help the model to dig deeper semantic information of the graph. In this paper, we utilize edge modification to create the second view. Then we sample subgraphs by random walk. The nodes and subgraphs will form the input to the graph contrastive network.

\subsubsection{Edge Modification.}
Edge Modification (EM) builds the second view by perturbing the graph edges. Inspired by~\cite{jin2021multi}, we not only delete edges in the adjacency matrix but also add the same number of edges simultaneously. In practice, we first set a fixed proportion $P$ to uniformly and randomly drop $\frac{PM}{2}$ edges from the adjacency matrix. Then $\frac{PM}{2}$ edges are added into the matrix uniformly and randomly. In this way, we attempt to learn robust representations of the subgraphs without destroying the properties of the graph. Ablation Study section will further discuss the graph augmentation method for generating the second view.

\subsubsection{Random Walk.}
For a target node, an effective anomaly detection method is measuring the feature distance between it and its neighborhoods~\cite{liu2021anomaly}. Thus, we employ random walk with restart (RWR)~\cite{qiu2020gcc} to sample the subgraphs around the nodes. A lower feature similarity indicates a higher anomalous degree of the target node.

\subsection{Graph Contrastive Network}
The contrastive learning paradigm has been proved effective for GAD~\cite{liu2021anomaly}. We construct a multi-view graph contrastive network, which includes three parts, i.e., node-subgraph, node-node, and subgraph-subgraph contrasts. The first two contrasts are defined in each view and will be strengthened by the information fusion of the two views. Node-subgraph contrast is mainly used to capture anomalous information from the neighborhood of nodes. The second contrast allows better detection of node-level anomalies. In the meantime, subgraph-subgraph contrast works between two views. It will straightforwardly optimize the subgraph embeddings for GAD, which significantly benefits node-subgraph contrast.

\subsubsection{Node-Subgraph Contrast.}
A target node $v_{i}$ forms a positive pair with its located subgraph and forms a negative pair with a random subgraph where another node $v_{j}$ is located. We first adopt a GCN layer that maps the features of nodes in the subgraph to the embedding space. It is worth noting that the features of the target node in the subgraph are masked, i.e., set to 0. The subgraph hidden-layer representation can be defined as:

\begin{equation}
\mathbf{H}^{\left (\ell+1  \right ) }_{i} =\sigma \left ( \mathbf{\widetilde{\mathbf{D}}}^{-\frac{1}{2}}_{i}\mathbf{\widetilde{\mathbf{A}}}_{i}\mathbf{\widetilde{\mathbf{D}}}^{-\frac{1}{2}}_{i}\mathbf{H}^{\left (\ell  \right ) }_{i}\mathbf{W}^{\left ( \ell \right ) }\right ),
\end{equation}
where $\mathbf{H}^{\left (\ell+1  \right ) }_{i}$ and $\mathbf{H}^{\left (\ell  \right ) }_{i}$ indicate the ${\left (\ell+1\right ) }$-th and ${\ell}$-th layer hidden representation, $\mathbf{\widetilde{\mathbf{D}}}^{-\frac{1}{2}}_{i}\mathbf{\widetilde{\mathbf{A}}}_{i}\mathbf{\widetilde{\mathbf{D}}}^{-\frac{1}{2}}_{i}$ is the normalization of the adjacency matrix, $\mathbf{W}^{\left ( \ell \right ) }$ denotes the network parameters.

Then, the subgraph final representation $\boldsymbol{z}_{i}$ is calculated by a \textit{Readout} function. Specifically, we utilize the average function to achieve \textit{Readout}:
\begin{equation}
\boldsymbol{z}_{i}=Readout\left (\mathbf{Z}_{i}\right ) =\sum_{j=1}^{n_{i}} \frac{\left (\mathbf{Z}_{i} \right )_{j}}{n_{i}}.
\end{equation}

Correspondingly, we leverage MLP to transform the target node features to the same embedding space as the subgraph. The node hidden-layer representation is:
\begin{equation}
\boldsymbol{h}^{\left (\ell+1  \right ) }_{i} =\sigma \left (\boldsymbol{h}^{\left (\ell  \right ) }_{i}\mathbf{W}^{\left ( \ell \right ) }\right ),
\end{equation}
where $\mathbf{W}^{\left ( \ell \right ) }$ is shared with the above GCN layer. $\boldsymbol{e}_{i}$ is the target node final embedding.

In each view, the anomalous degree of the target node is related to the similarity $s_{i}^{1}$ of the subgraph and the node embeddings. We adopt a \textit{Bilinear} model to measure the relationship:
\begin{equation}
s_{i}^{1} =Bilinear\left (\boldsymbol{z}_{i},\boldsymbol{e}_{i}\right ) =sigmoid \left (\boldsymbol{z}_{i}\mathbf{W}\boldsymbol{e}_{i}^\top\right ).
\label{node_subgraph_similarity}
\end{equation}

Generally, the target node and subgraph representations tend to be similar in positive pairs, i.e., $s_{i}^{1}=1$. On the contrary, they may be dissimilar in negative pairs, i.e., $s_{i}^{1}=0$. Hence, we employ the binary cross-entropy (BCE) loss~\cite{velickovic2019deep} to train the contrast:
\begin{equation}
\mathcal{L}_{NS}^{1}=- \sum_{i=1}^{N}\left (y_{i}\log{\left (s_{i}^{1} \right )} +  \left (1 - y_{i}\right )\log{\left ( 1 - s_{i}^{1} \right ) }\right ),
\label{node_subgraph_loss1}
\end{equation}
where $y_{i}$ is equal to 1 in positive pairs, and is equal to 0 in negative pairs.

We can obtain the similarity degree $s_{i}^{2}$ and BCE loss $\mathcal{L}_{NS}^{2}$ in another view as well. It is worth mentioning that the two networks under two views use the same architecture and share parameters. Therefore, the final node-subgraph contrast loss is:

\begin{equation}
\mathcal{L}_{NS}= \alpha \mathcal{L}_{NS}^{1}+\left ( 1-\alpha  \right ) \mathcal{L}_{NS}^{2},
\label{node_subgraph_loss}
\end{equation}
where ${\alpha \in \left ( 0,1 \right ) }$ is a trade-off parameter to balance the importance between two views.

\subsubsection{Node-Node Contrast.}
Node-node contrast can effectively discover the node-level anomalies. Similarly, the target node features will be masked. And its representation is aggregated from another nodes in the subgraph. In each view, it forms a positive pair with the same node after MLP, and forms a negative pair with another node after MLP. We leverage a new GCN to obtain the representation of the subgraph:
\begin{equation}
\mathbf{H}^{\prime\left (\ell+1  \right ) }_{i} =\sigma \left ( \mathbf{\widetilde{\mathbf{D}^{\prime}}}^{-\frac{1}{2}}_{i}\mathbf{\widetilde{\mathbf{A}^{\prime}}}_{i}\mathbf{\widetilde{\mathbf{D}^{\prime}}}^{-\frac{1}{2}}_{i}\mathbf{H}^{\prime\left (\ell  \right ) }_{i}\mathbf{W}^{\prime\left ( \ell \right ) }\right ),
\label{node_GCN}
\end{equation}
where $\mathbf{W}^{\prime\left ( \ell \right ) }$ is different from parameter matrix used in node-subgraph contrast. $\boldsymbol{h}_{i}^{\prime\left (\ell+1  \right ) }=\mathbf{H}^{\prime\left (\ell+1  \right ) }_{i}\left [ 1,: \right ]$ is the ${\left (\ell+1\right ) }$-th layer hidden representation of $v_{i}$. And $\boldsymbol{u}_{i}$ is the target node final embedding.

In the meantime, we use a MLP to map the node features into the same hidden space:
\begin{equation}
\boldsymbol{\hat{h}}^{\left (\ell+1  \right ) }_{i} =\sigma \left (\boldsymbol{\hat{h}}^{\left (\ell  \right ) }_{i}\mathbf{W}^{\prime\left ( \ell \right ) }\right ),
\end{equation}
where $\mathbf{W}^{\prime\left ( \ell \right ) }$ is shared with Eq.~\eqref{node_GCN}. After MLP, $\boldsymbol{\hat{e}}_{i}$ is the target node final embedding.

Similar to the node-subgraph contrast, we adopt a \textit{Bilinear} model to evaluate the relationship $\hat{s}^{1}_{i}$ between $\boldsymbol{u}_{i}$ and $\boldsymbol{\hat{e}}_{i}$. Then the node-node contrast loss function can be defined as:

\begin{equation}
\mathcal{L}_{NN}^{1}=- \sum_{i=1}^{N}\left (\hat{y}_{i}\log{\left (\hat{s}^{1}_{i} \right )} +  \left (1 - \hat{y}_{i}\right )\log{\left ( 1 - \hat{s}^{1}_{i} \right ) }\right ).
\label{node_node_loss1}
\end{equation}

Likewise, similarity degree $\hat{s}^{2}_{i}$ and loss $\mathcal{L}_{NN}^{2}$ can also be computed for another view. So the final node-node contrast loss function is:

\begin{equation}
\mathcal{L}_{NN}= \alpha \mathcal{L}_{NN}^{1}+\left ( 1-\alpha  \right ) \mathcal{L}_{NN}^{2},
\label{node_node_loss}
\end{equation}
where the view-balance parameter $\alpha$ is shared with the node-subgraph contrast loss in Eq.~\eqref{node_subgraph_loss}.

\subsubsection{Subgraph-Subgraph Contrast.}
Subgraph-subgraph contrast is defined between two views. It aims to learn more representative and intrinsic subgraph embeddings for GAD, which will help node-subgraph contrast discriminate the relationship of nodes and their neighborhoods. In practice, we directly optimize the subgraph representations under the joint loss with node-subgraph contrast.

A subgraph forms a positive pair with the perturbed subgraph where the same target node $v_{i}$ locates in another view. Different from common graph contrastive methods~\cite{you2020graph}, it forms negative pairs with two subgraphs where another node $v_{j}$ locates in two views. The node $v_{j}$ is the same one which subgraph forms a negative pair with $v_{i}$ in node-subgraph contrast. Inspired by~\cite{oord2018representation}, we employ a loss function to optimize the contrast:

\begin{equation}
  \mathcal{L}_{SS}=-\sum_{i=1}^{n}\log{\frac{\exp{\left(\boldsymbol{z}^{1}_i\cdot\boldsymbol{z}_{i}^{2}\right)}}{\exp{\left( \boldsymbol{z}_i^{1}\cdot\boldsymbol{z}_j^{1}\right)}+\exp{\left( \boldsymbol{z}_i^{1}\cdot\boldsymbol{z}_j^{2}\right)}}},
  \label{subgraph_subgraph_loss}
\end{equation}
where $\boldsymbol{z}^{1}_i$ and $\boldsymbol{z}^{2}_i$ represent the embeddings of the subgraphs that the target node $v_{i}$ belongs to in two views. And $\boldsymbol{z}^{1}_j$ and $\boldsymbol{z}^{2}_j$ are the embeddings of the subgraphs where another node $v_{j}$ locates in two views, respectively.

\subsubsection{Loss Function.}
To integrate the advantages of three contrasts, we optimize the joint loss function:

\begin{equation}
\mathcal{L}= \beta \mathcal{L}_{NS}+\left ( 1-\beta  \right ) \mathcal{L}_{NN}+\gamma\mathcal{L}_{SS},
\label{total_loss}
\end{equation}
where $\beta \in \left ( 0,1 \right )$ is a balance parameter of subgraph-level and node-level anomalous information. $\gamma\in \left ( 0,1 \right )$ is a trade-off parameter of $\mathcal{L}_{SS}$.

\begin{algorithm}[!t]
\small
\caption{Proposed model GRADATE.}
\label{ALGORITHM}
\flushleft{\textbf{Input}: An undirected graph $\mathcal{G} = \left ( \mathcal{V}, \mathcal{E} \right )$; Number of training epochs $E$; Batch size $B$.} \\
\flushleft{\textbf{Output}: Anomaly score function $f$.} 
\begin{algorithmic}[1]
\FOR{$e=1$ to $E$}
\STATE Form two views from the original graph and the graph augmented by edge modification. And sample subgraphs by random walk in each view. 
\STATE $\mathcal{V}$ is divided into batches with size $B$ by random.
\FOR{$v_{i} \in B$}
\STATE In node-subgraph contrast, estimate the representation similarity of the target node and two subgraphs in positive and negative pairs via Eq.~\eqref{node_subgraph_similarity}. 
\STATE In node-node contrast, calculate the representation similarity of the target node and two nodes in positive and negative pairs.
\STATE Calculate the node-subgraph, node-node, and subgraph-subgraph loss via Eq.~\eqref{node_subgraph_loss},~\eqref{node_node_loss} and~\eqref{subgraph_subgraph_loss}. Then obtain the final joint loss via Eq.~\eqref{total_loss}.
\STATE Back propagation and update trainable parameters.
\ENDFOR
\ENDFOR
\STATE By multi-round detections, calculate the final anomaly score for each node via 
Eq.~\eqref{anomaly_score1},~\eqref{anomaly_score2} and~\eqref{anomaly_score3}.
\end{algorithmic}
\end{algorithm}

\subsubsection{Anomaly Score Calculation.}
In node-subgraph and node-node contrasts, a normal node is similar to the subgraph or node in its positive pair and dissimilar to the subgraph or node in its negative pair. On the contrary, an anomalous node is dissimilar to the node or subgraph in both positive and negative pairs. Naturally, we define the anomaly score of the target node as follows:

\begin{equation}
s_{i} = s_{i}^{n} - s_{i}^{p},
\label{anomaly_score1}
\end{equation}
where $s_{i}^{p}$ and $s_{i}^{n}$ represent the similarity of positive and negative pairs. 

Then we comprehensively fuse the anomalous information from two views and three contrast. The anomaly score can further be represented as:
\begin{equation}
\begin{split}
s_{i}&=\alpha s_{i}^{1} +\left ( 1-\alpha  \right ) s_{i}^{2},\\
\hat{s}_{i}&=\alpha \hat{s}_{i}^{1} +\left ( 1-\alpha  \right ) \hat{s}_{i}^{2},\\
S_{i}&=\beta s_{i} +\left ( 1-\beta  \right ) \hat{s}_{i},
\end{split}
\label{anomaly_score2}
\end{equation}
where $\alpha$ and $\beta$ are shared with Eq.~\eqref{node_subgraph_loss},~\eqref{node_node_loss} and~\eqref{total_loss}.

Once detection with only once random walk cannot capture sufficient semantic information. Multi-round detections are essential to compute anomaly score for each node. Inspired by~\cite{jin2021anemone}, we calculate the final anomaly score by the mean and the standard deviation from the results of multi-round detections:
\begin{equation}
\begin{split}
\bar{S}_{i}&=\frac{1}{R}\sum_{r=1}^{R}S_{i}^{\left (r \right )}, \\
S_{i}&=\bar{S}_{i}+\sqrt{\frac{1}{R}\sum_{r=1}^{R}\left (S_{i}^{\left (r \right )}-\bar{S}_{i}\right )^{2}  },
\end{split}
\label{anomaly_score3}
\end{equation}
where $R$ is the number of anomaly detection rounds.

In general, the overall procedures of GRADATE are shown in Algorithm~\ref{ALGORITHM}.

\section{Experiments}
We conduct extensive experiments on six graph benchmark datasets to verify the excellent performance of GRADARE. The results also confirm the effectiveness of subgraph-subgraph contrast and edge modification for GAD.

\subsection{Experiment Settings}
Details of the experiment settings are shown as follows: \textbf{(1)Datasets.} The proposed method is evaluated on six benchmark datasets which details are shown in Table~\ref{table:datasets}. The datasets include Citation~\cite{yuan2021higher}, Cora~\cite{sen2008collective}, WebKB~\cite{craven1998learning}, UAI2010~\cite{wang2018unified}, UAT and EAT~\cite{mrabah2022rethinking}. \textbf{(2)Anomaly Injection.} Following DOMINANT, we inject the same number of feature and structure anomalies into the original datasets which do not have anomalous nodes before. The total number of anomalies for each dataset is shown in the last column of Table~\ref{table:datasets}. \textbf{(3)Baselines.} For the GAD task, we compare with eight well-known baseline methods. They are summarized in the first column of Table~\ref{table:AUC}. The first two models are non-deep algorithms, and the rest are based on graph neural networks. Following CoLA, the node features of all datasets are reduced to 30 by PCA before running ANOMALOUS. \textbf{(4)Metric.} We adopt a widely-used anomaly detection metric AUC to evaluate the above methods.

\begin{table}[h]
\centering
\caption{The statistics of datasets.}
\resizebox{0.45\textwidth}{15mm}{
\begin{tabular}{c|cccc}
\toprule
$\textbf{Datasets}$&$\textbf{Nodes}$&$\textbf{Edges}$&$\textbf{Attributes}$&$\textbf{Anomalies}$\\
\midrule
$\textbf{EAT}$& 399  & 5993 & 203       & 30       \\
$\textbf{WebKB}$& 919  & 1662 & 1703       & 60       \\
$\textbf{UAT}$& 1190  & 13599  & 239       & 60       \\
$\textbf{Cora}$& 2708  & 5429   & 1433       & 150       \\
$\textbf{UAI2010}$& 3067 & 28311  & 4973        & 150       \\
$\textbf{Citation}$& 8935  & 15098 & 6775       & 450       \\
\bottomrule
\end{tabular}}
\label{table:datasets}
\end{table}


\begin{figure*}[t]
\centering
\subfigure[EAT]{
\includegraphics[width=0.31\linewidth]{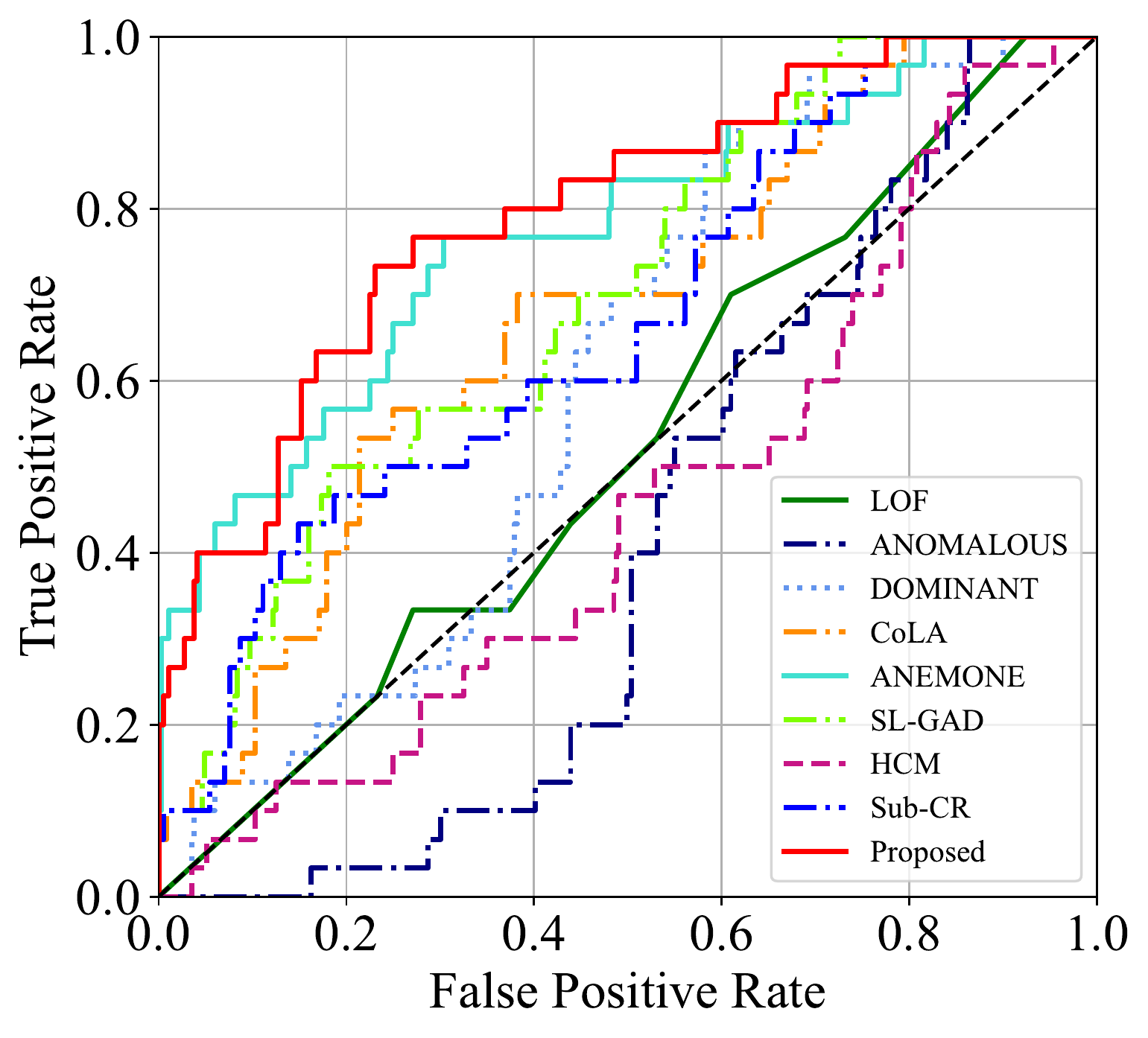}
}
\subfigure[WebKB]{
\includegraphics[width=0.31\linewidth]{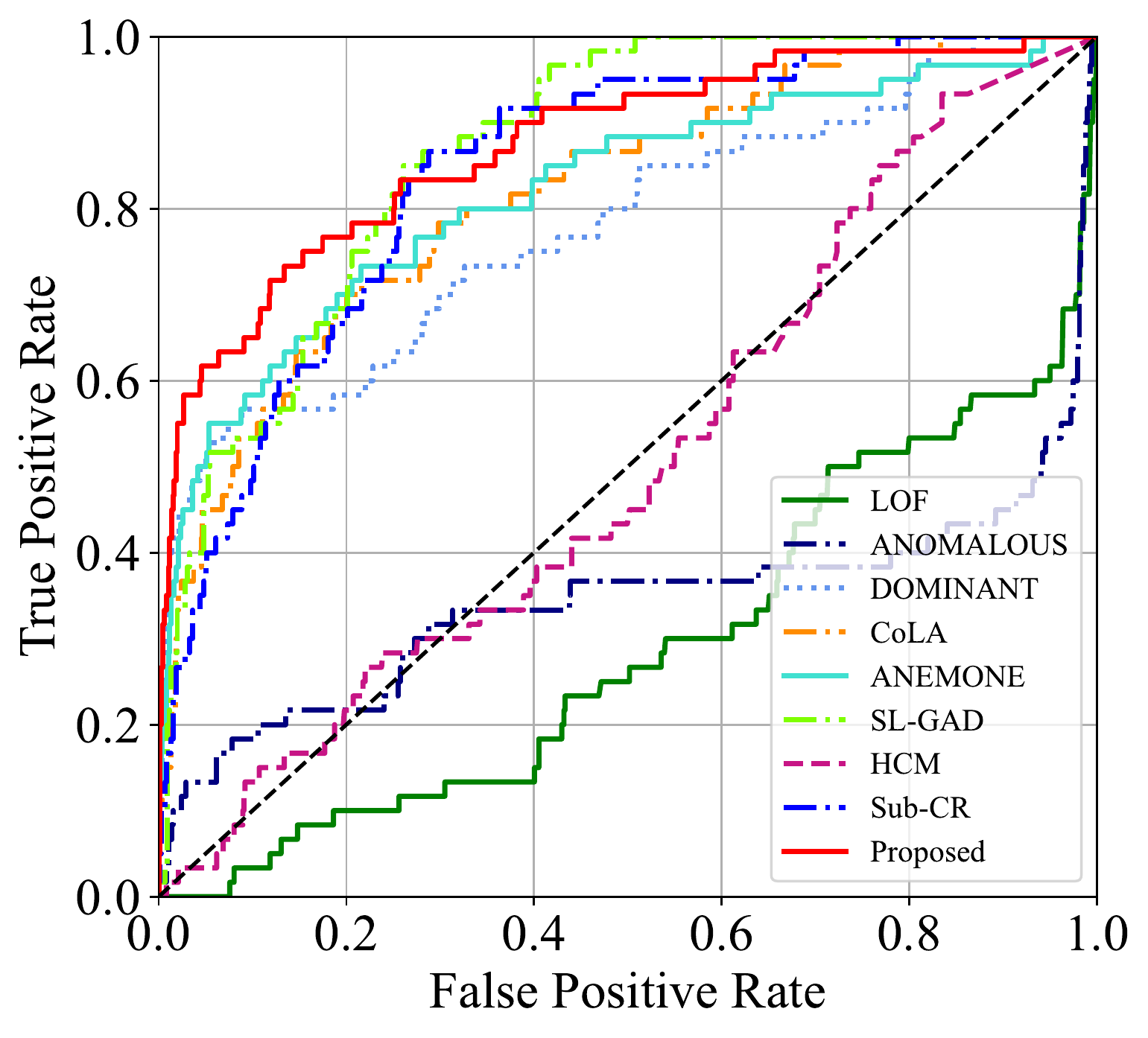}
}
\subfigure[UAT]{
\includegraphics[width=0.31\linewidth]{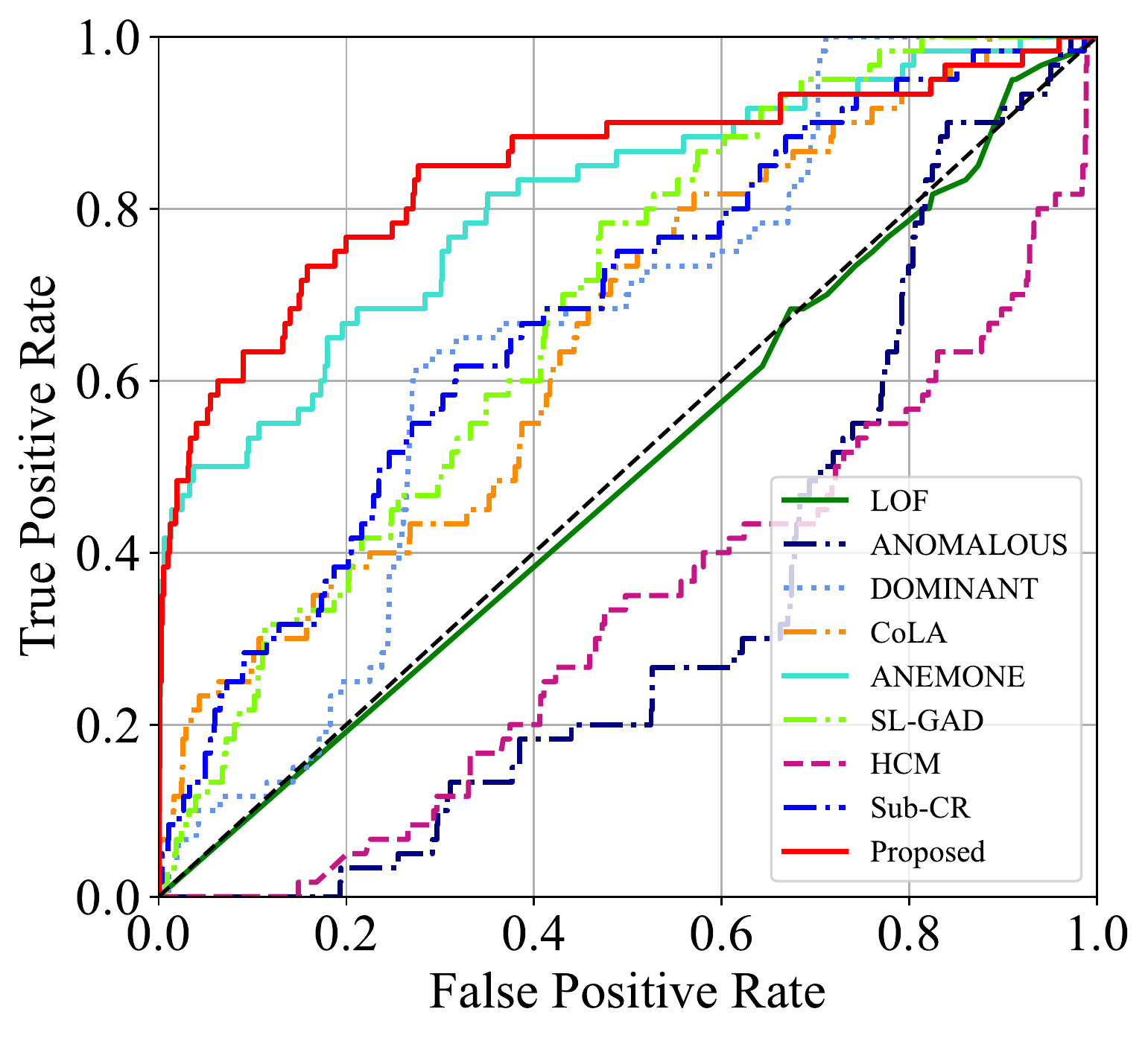}
}
\subfigure[Cora]{
\includegraphics[width=0.31\linewidth]{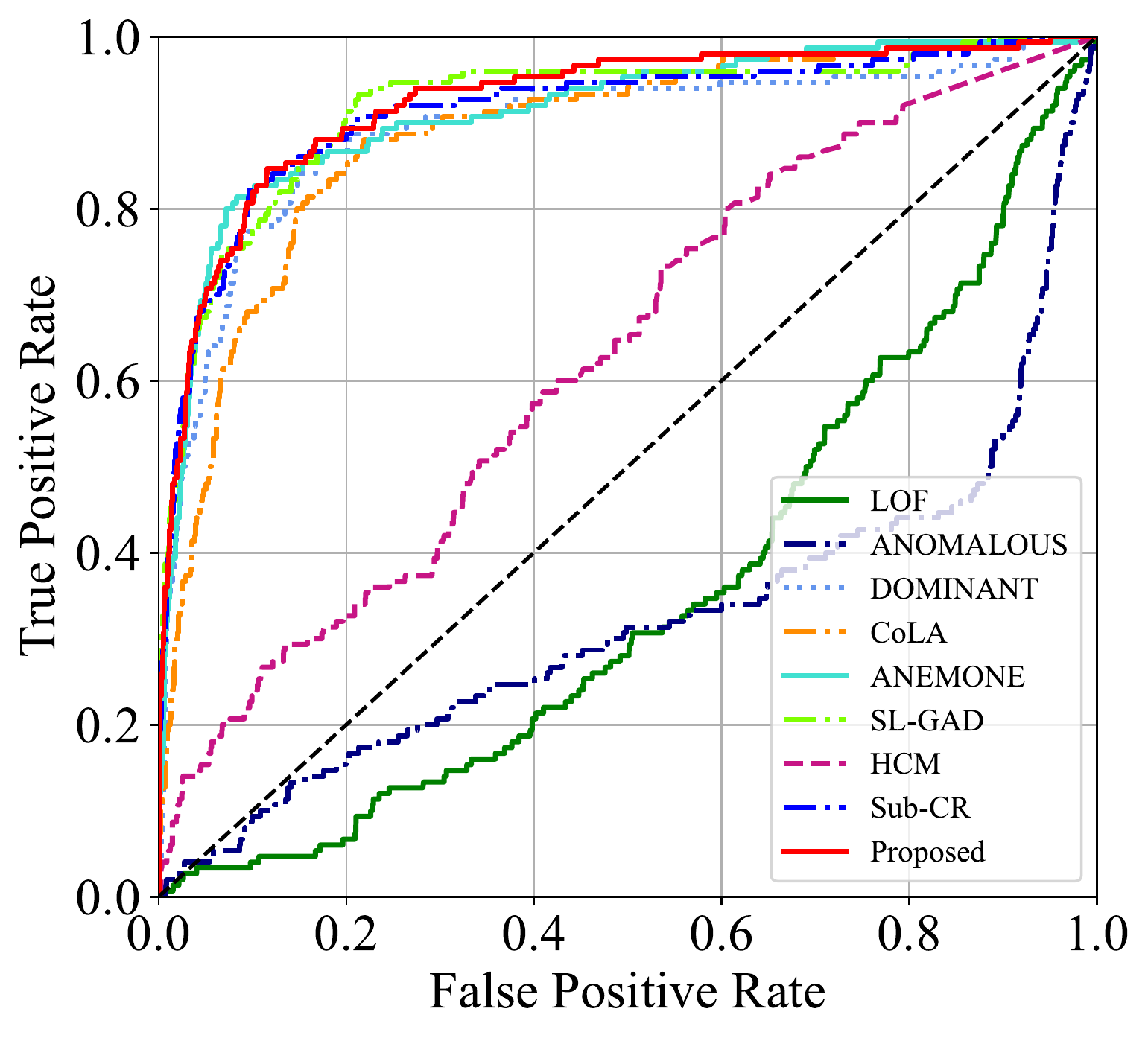}
}
\subfigure[UAI2010]{
\includegraphics[width=0.31\linewidth]{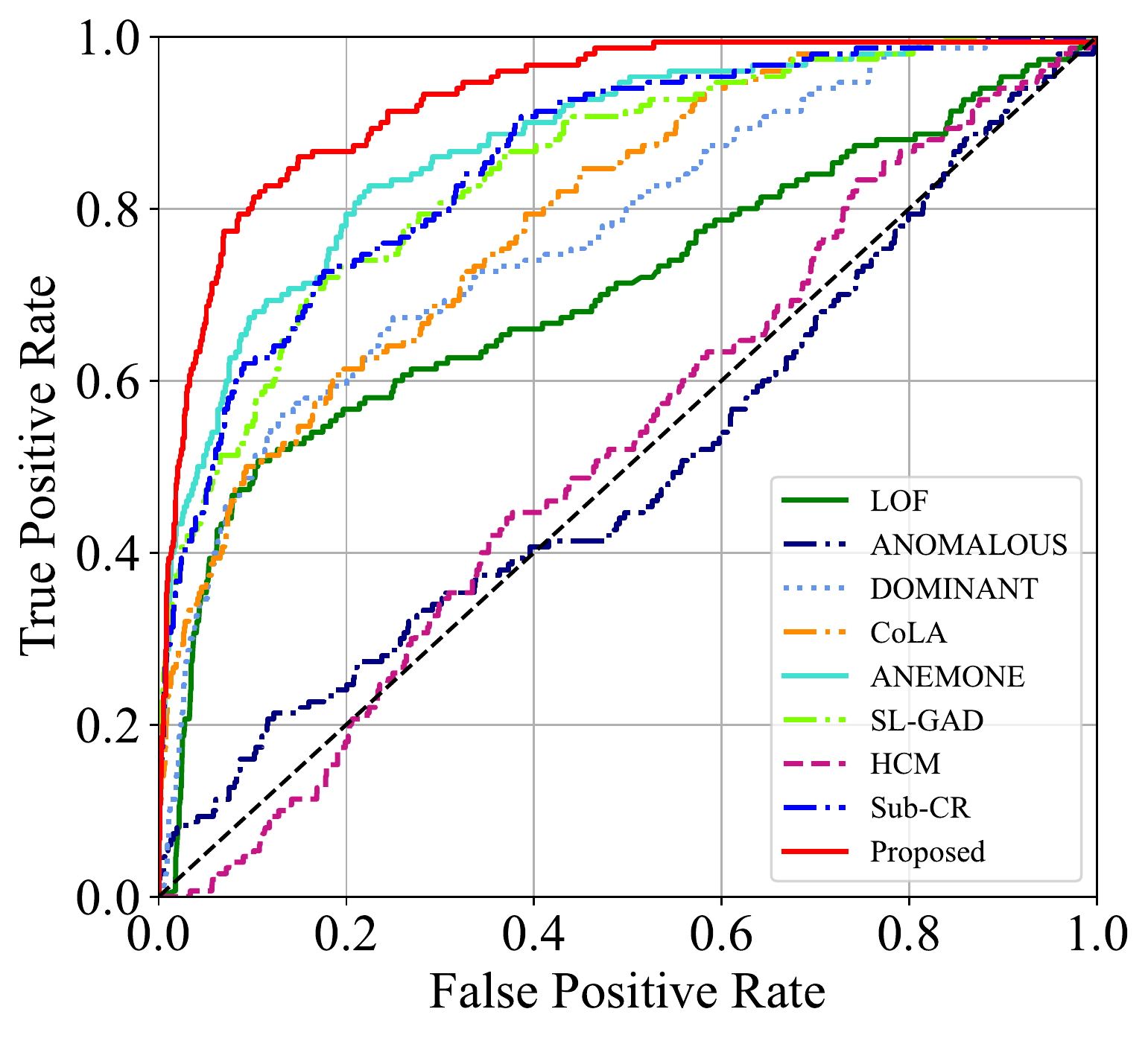}
}
\subfigure[Citation]{
\includegraphics[width=0.31\linewidth]{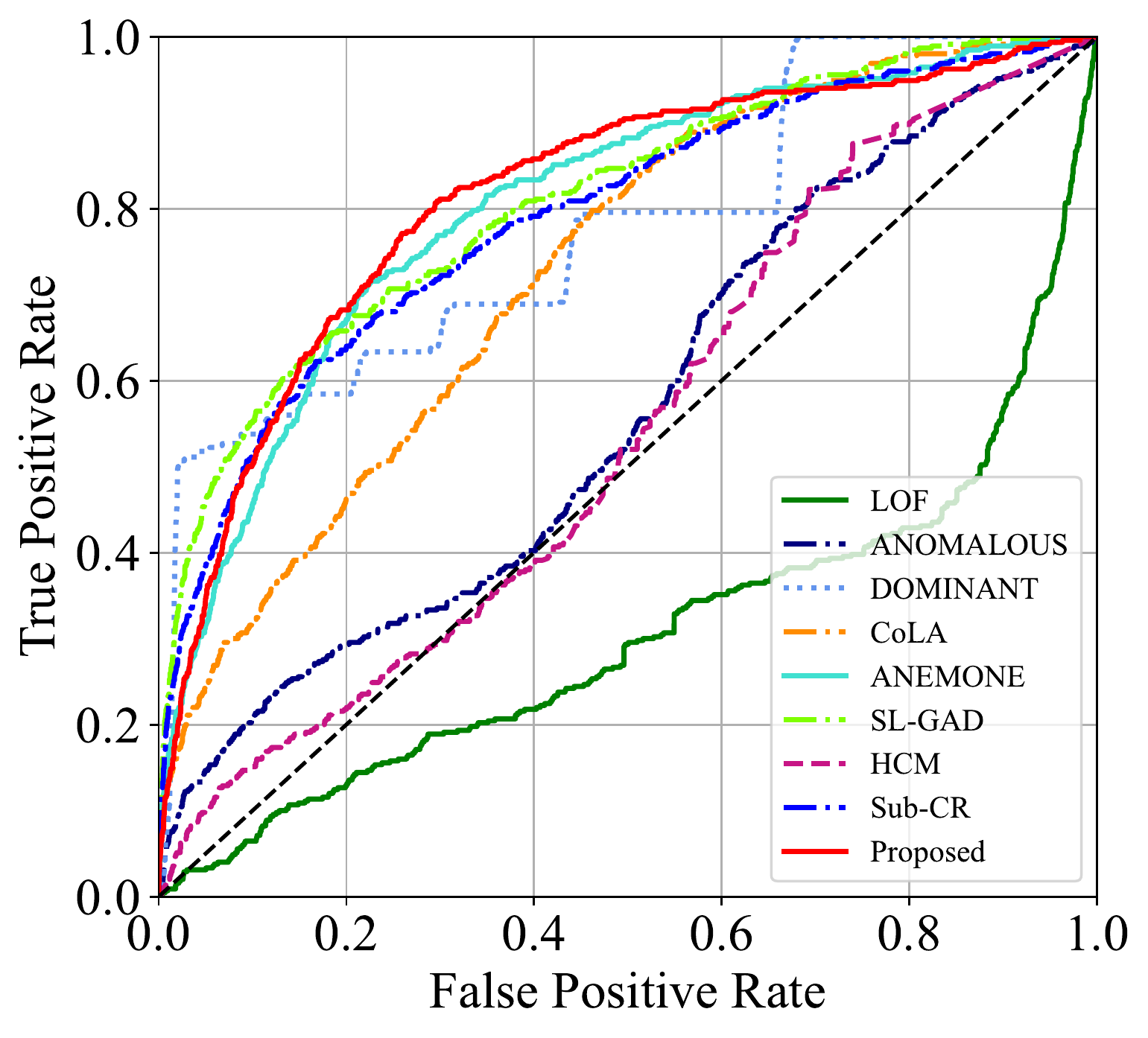}
}
\vspace{-10pt}
\caption{ROC curves on six benchmark datasets. The area under the curve is larger, the anomaly detection performance is better. The black dotted lines are the ``random line'', indicating the performance under random guessing.}
\label{fig:ROC}
\vspace{-5pt}
\end{figure*}

\begin{table*}[!ht]
\centering
\small
\setlength{\tabcolsep}{5.0mm}
\caption{Performance comparison for AUC. The bold and underlined values indicate the best and runner-up results, respectively.}
\vspace{-5pt}
\resizebox{0.9\textwidth}{!}{
\begin{tabular}{ccccccc}
\toprule
 \textbf{Methods} & \textbf{EAT} & \textbf{WebKB}   & \textbf{UAT} & \textbf{Cora}    & \textbf{UAI2010}  & \textbf{Citation} \\ \midrule
 LOF~\cite{breunig2000lof} & 0.5255      & 0.2903  & 0.4906   & 0.3538          & 0.7052                              & 0.3059                           \\
 ANOMALOUS~\cite{peng2018anomalous} & 0.4109      & 0.3417   & 0.3356    & 0.3198          & 0.5026                           & 0.5656                    \\
 DOMINANT~\cite{ding2019deep} & 0.6023       & 0.7787 & 0.6503   & 0.8929          & 0.7698                             & 0.7748                           \\
 CoLA~\cite{liu2021anomaly} & 0.6762       & 0.8175   & 0.6538  & 0.8847          & 0.7949                            & 0.7296                           \\
 ANEMONE~\cite{jin2021anemone} & \underline{0.7726}    & 0.8208   & \underline{0.8087}  & 0.9122          & \underline{0.8731}                 & 0.8028     \\
 SL-GAD~\cite{zheng2021generative} & 0.6974    & \underline{0.8678} & 0.6851     & \underline{0.9192}    & 0.8454                    & \underline{0.8095}        \\
 HCM~\cite{huang2021hop} & 0.4536       & 0.5064  & 0.3262    & 0.6276          & 0.5210                          & 0.5414                    \\
 Sub-CR~\cite{zhang2022reconstruction} & 0.6672       & 0.8423   & 0.6788       & 0.9133          & 0.8571                        &0.7903               \\\midrule
 \textbf{GRADATE}    & \textbf{0.7980}  & \textbf{0.8740} & \textbf{0.8451} & \textbf{0.9237} & \textbf{0.9262}    & \textbf{0.8138}                  \\ \bottomrule
\end{tabular}}
\label{table:AUC}
\vspace{-10pt}
\end{table*}

\subsection{Model Parameters}
In node-subgraph and subgraph-subgraph contrasts, both GCN models have one layer and use ReLU as the activation function. The size of subgraphs in the network is set to 4. Both node and subgraph features are mapped to 64 dimensions in hidden space. Besides, we implement 400 epochs of model training and 256 rounds of anomaly score calculation. 

\subsection{Result and Analysis}
In this subsection, we evaluate the anomaly detection performance of GRADATE by comparing it with eight baseline methods. Figure~\ref{fig:ROC} demonstrates the ROC curves for nine models. Meanwhile, Table~\ref{table:AUC} shows the comparison results of AUC values corresponding to Figure~\ref{fig:ROC}. For the results, we have the following conclusions:

\begin{itemize}
    \item We can intuitively find that GRADATE outperforms its competitors on these six datasets. To be specific, GRADATE achieves notable AUC gains of \textbf{2.54\%}, \textbf{0.62\%}, \textbf{3.64\%}, \textbf{0.45\%}, \textbf{5.31\%}, and \textbf{0.43\%} on EAT, WebKB, UAT, Cora, UAI2010 and Citation, respectively. And as shown in Figure~\ref{fig:ROC}, the under-curve areas of GRADATE are significantly larger than competitors.
    \item We observe that most neural network-based methods outperform the shallow methods, LOF and ANOMALOUS. Shallow methods have inherent limitations on dealing with the high-dimension features of graph data. 
    \item Among the deep methods, contrastive learning-based methods, CoLA, ANEMONE, SL-GAD, Sub-CR, and GRADATE work better. It indicates that the contrastive learning-based pattern can effectively detect anomalies by mining the feature and structure information from graph. With the newly added subgraph-subgraph contrast and multi-view learning strategies, GRADATE achieves the best performance.
\end{itemize}

\subsection{Ablation Study}
\label{ablation}

\subsubsection{Various Scale Contrastive Strategy.} To verify the effectiveness of the proposed subgraph-subgraph contrast, we perform ablation study experiments. For convenience, \textbf{NS}, \textbf{NS+SS}, \textbf{NS+NN}, and \textbf{NS+NN+SS} indicate only using node-subgraph contrast (CoLA), using node-subgraph and subgraph-subgraph contrasts, using node-subgraph and node-node contrasts (ANEMONE), and using above three contrasts (GRADATE), separately. As shown in Table~\ref{table:contrast}, adding subgraph-subgraph contrast can enhance the detection performance by boosting node-subgraph contrast. Using all three contrasts will lead to the best performance.

\begin{table}[ht]
\centering
\caption{Ablation study for contrast scale w.r.t. AUC.}
\resizebox{0.47\textwidth}{!}{
\begin{tabular}{ccccccc}
\toprule
&$\textbf{EAT}$&$\textbf{WebKB}$&$\textbf{UAT}$&$\textbf{Cora}$&$\textbf{UAI2010}$&$\textbf{Citation}$\\
\midrule
NS& 0.6762 & 0.7949  & 0.6538 & 0.8847 & 0.8175 & 0.7296  \\
NS+SS& 0.6800 & 0.8310 & 0.6603  & 0.8956 & 0.9055   & 0.6978  \\
\midrule
NS+NN& 0.7726 & 0.8208 & 0.8087  & 0.9122  & 0.8731  & 0.8028  \\
\textbf{NS+NN+SS}& \textbf{0.7980} & \textbf{0.8740}& \textbf{0.8451}   & \textbf{0.9237}  & \textbf{0.9262}  & \textbf{0.8138}  \\
\bottomrule
\end{tabular}}
\label{table:contrast}
\end{table}

\subsubsection{Graph Augmentation Strategy.} In the meantime, we adopt four different graph augmentation to form the second view and explore their effects on performance. \textbf{Gaussian Noised Feature (GNF)} means node features are randomly perturbed with Gaussian noise. \textbf{Feature Mask (FM)} indicates random parts of node features are masked. \textbf{Graph Diffusion (GD)} is that the graph diffusion matrix is generate by diffusion model~\cite{hassani2020contrastive, klicpera2019diffusion}. GNF and FM are perturbations of node features. GD and \textbf{Edge Modification (EM)} are widely-used graph augmentation methods on graph edges. As shown in Table~\ref{table:augmentation}, EM achieves the best performance on all datasets. On further analysis, perturbation of node features may disrupt the features of normal nodes. It will harm the comparison between the nodes and their neighborhoods, which is the basis of contrastive learning for GAD. This can cause some normal nodes to be misclassified as anomalies and lead to performance degradation. Moreover, GD is a structure-based graph augmentation method. However, its primary purpose is to capture global information. Hence, EM is more compatible with subgraph-subgraph contrast than GD, which can improve node-subgraph contrast to dig the local neighborhood information of nodes.

\begin{table}[ht]
\centering
\caption{Ablation study for graph augmentation w.r.t. AUC.}
\resizebox{0.47\textwidth}{!}{
\begin{tabular}{ccccccc}
\toprule
 &$\textbf{EAT}$&$\textbf{WebKB}$&$\textbf{UAT}$&$\textbf{Cora}$&$\textbf{UAI2010}$&$\textbf{Citation}$\\
\midrule
GNF& 0.7548 & 0.8183 & 0.8327  & 0.9031 & 0.9193  & 0.7902  \\
FM & 0.7782  & 0.8148  & 0.8256  & 0.8924  & 0.9171& 0.8034 \\
GD& 0.7618 & 0.8062  & 0.8143   & 0.9026& 0.9161  & 0.8030  \\
\textbf{EM} & \textbf{0.7980}  & \textbf{0.8740} & \textbf{0.8451}  & \textbf{0.9237}  & \textbf{0.9262}  & \textbf{0.8138}    \\
\bottomrule
\end{tabular}}
\label{table:augmentation}
\end{table}

\subsection{Sensitivity Analysis}
\subsubsection{Balance Parameter $\alpha$, $\beta$ and $\gamma$.} 
\begin{figure}[!htbp]
\centering
\subfigure[EAT]{
\includegraphics[width=0.45\linewidth]{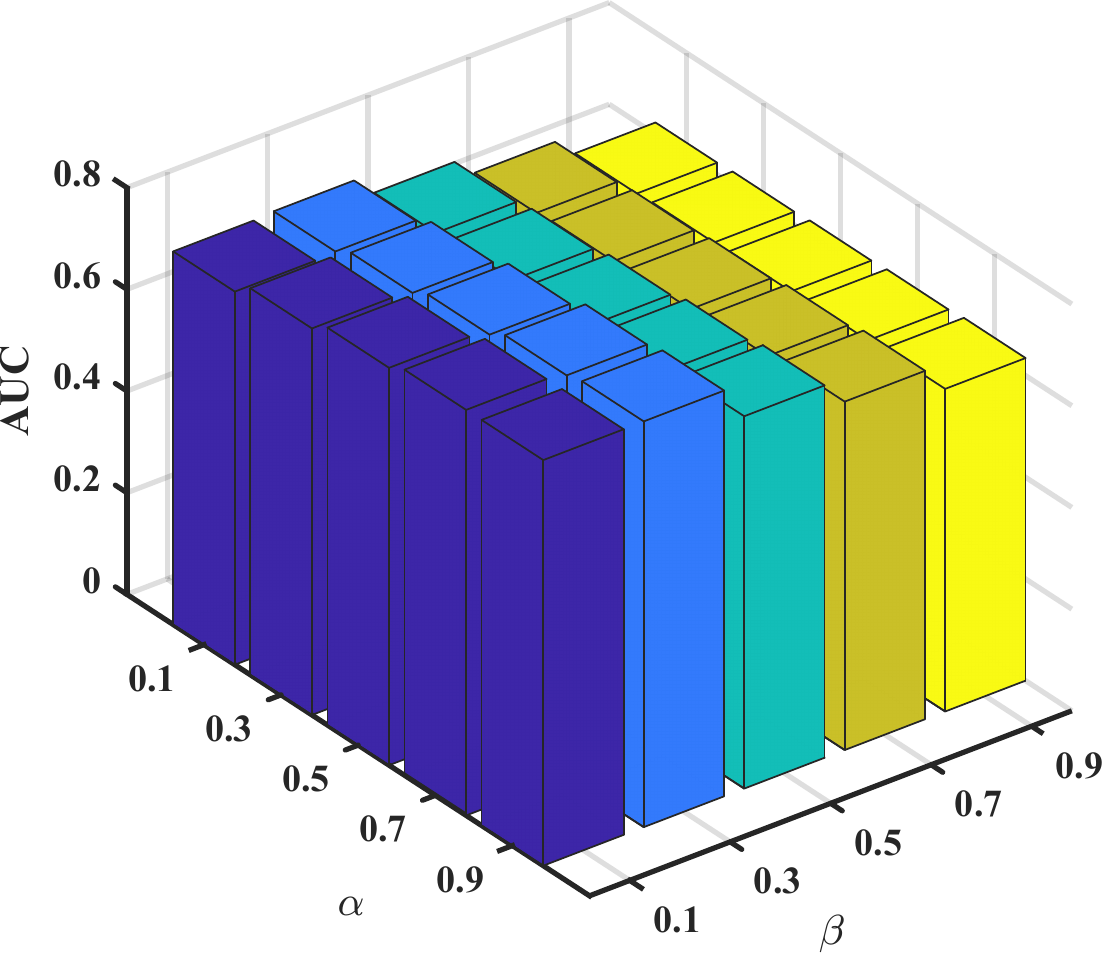}
}
\subfigure[UAI2010]{
\includegraphics[width=0.45\linewidth]{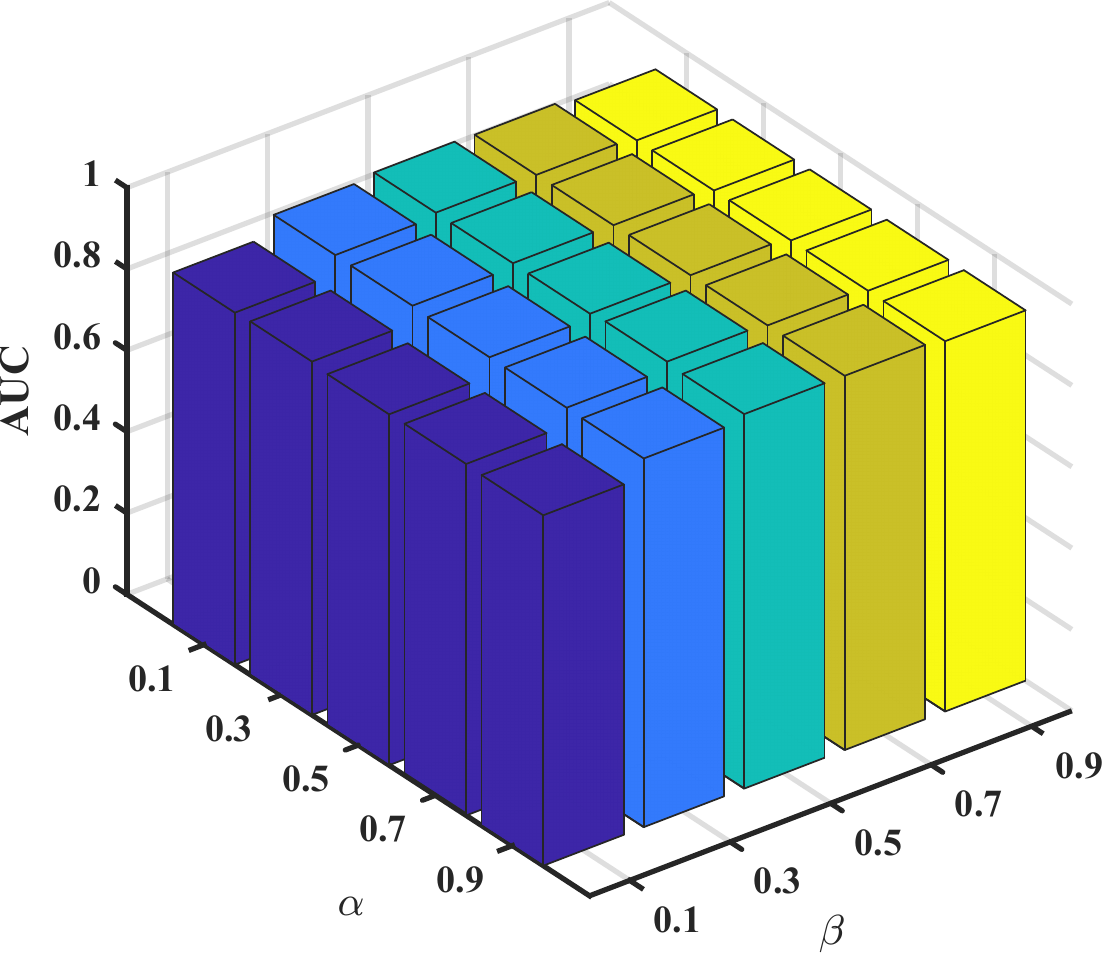}
}
\caption{Sensitivity analysis for the balance parameters $\alpha$ and $\beta$ w.r.t. AUC on EAT and UAI2010.}
\label{fig:alpha_beta}
\end{figure}

We discuss the three important balance parameters in the loss function. As shown in Figure~\ref{fig:alpha_beta}, the hyper-parameter $\alpha$ and $\beta$ are effective in improving the detection performance on EAT and UAI2010. Similar phenomena can be observed on the other datasets. In practice, we set $\alpha$ to 0.9, 0.1, 0.7, 0.9, 0.7, and 0.5 on EAT, WebKB, UAT, Cora, UAI2010 and Citation. Meanwhile, we make $\beta$ to 0.3, 0.7, 0.1, 0.3, 0.5, and 0.5. 

\begin{figure}[ht]
    \centering
    \includegraphics[width = 0.40\textwidth]{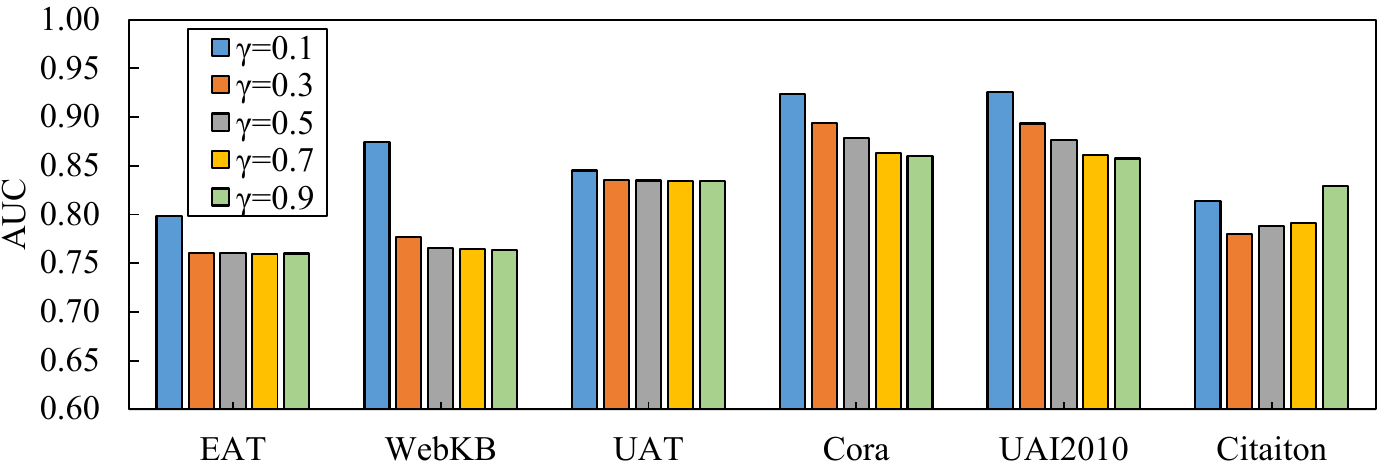}
    \caption{Trade-off parameter ${\gamma}$ w.r.t. AUC.}
    \label{fig:Gamma}
\end{figure}

Figure~\ref{fig:Gamma} illustrates the performance variation of GRADATE when $\gamma$ varies from 0.1 to 0.9. From the figure, we observe that GRADATE tends to perform well by setting $\gamma$ to 0.1 across all benchmarks.

\subsubsection{Edge Modification Proportion $P$.}
\begin{figure}[ht]
    \centering
    \includegraphics[width = 0.40\textwidth]{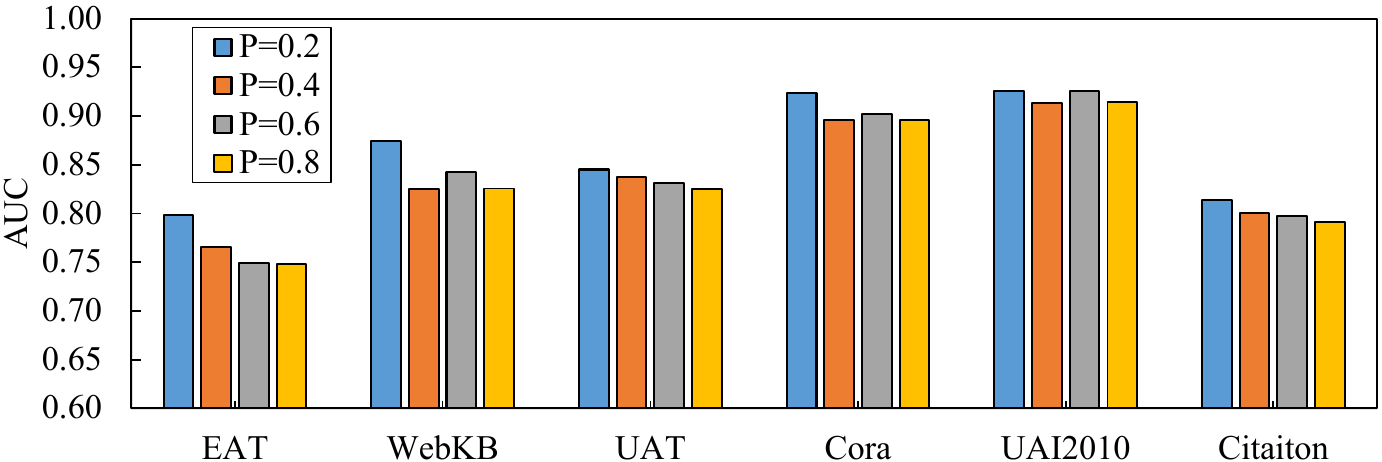}
    \caption{Perturbation proportion ${P}$ w.r.t. AUC.}
    \label{fig:P}
\end{figure}

We also investigate the influence of different parameterized edge modification. Figure~\ref{fig:P} shows that detection performance receives a relatively small fluctuation by modification proportion $P$ on UAT, UAI2010, and Citation. Comprehensively, we fixedly set $P=0.2$ on all datasets.

\section{Conclusion}
In this paper, we propose a new graph anomaly detection framework via multi-scale contrastive learning networks with an augmented view. We introduce subgraph-subgraph contrast to GAD and investigate the impact of different graph augmentation technologies on the task. Extensive experiments on six benchmark datasets prove GRADATE outperforms the competitors. In the future, we will continue to explore the contrastive learning pattern for the task.

\section{Acknowledgments}
This work was supported by the National Key R\&D Program of China (project no. 2020AAA0107100) and the National Natural Science Foundation of China (project no. 61922088, 61976196 and 61872377).

\bigskip

\bibliography{GRADATE}

\begin{thebibliography}{57}
\providecommand{\natexlab}[1]{#1}

\bibitem[{Breunig et~al.(2000)Breunig, Kriegel, Ng, and
  Sander}]{breunig2000lof}
Breunig, M.~M.; Kriegel, H.-P.; Ng, R.~T.; and Sander, J. 2000.
\newblock LOF: identifying density-based local outliers.
\newblock In \emph{Proc. of SIGMOD}.

\bibitem[{Cheng et~al.(2021{\natexlab{a}})Cheng, Wang, Zhang, Wang, Liu, and
  Zhu}]{cheng2021improved}
Cheng, Z.; Wang, S.; Zhang, P.; Wang, S.; Liu, X.; and Zhu, E.
  2021{\natexlab{a}}.
\newblock Improved autoencoder for unsupervised anomaly detection.
\newblock \emph{International Journal of Intelligent Systems}.

\bibitem[{Cheng et~al.(2021{\natexlab{b}})Cheng, Zhu, Wang, Zhang, and
  Li}]{cheng2021unsupervised}
Cheng, Z.; Zhu, E.; Wang, S.; Zhang, P.; and Li, W. 2021{\natexlab{b}}.
\newblock Unsupervised outlier detection via transformation invariant
  autoencoder.
\newblock \emph{IEEE Access}.

\bibitem[{Craven et~al.(1998)Craven, DiPasquo, Freitag, McCallum, Mitchell,
  Nigam, and Slattery}]{craven1998learning}
Craven, M.; DiPasquo, D.; Freitag, D.; McCallum, A.; Mitchell, T.~M.; Nigam,
  K.; and Slattery, S. 1998.
\newblock Learning to Extract Symbolic Knowledge from the World Wide Web.
\newblock In \emph{Proc. of AAAI}.

\bibitem[{Ding et~al.(2019)Ding, Li, Bhanushali, and Liu}]{ding2019deep}
Ding, K.; Li, J.; Bhanushali, R.; and Liu, H. 2019.
\newblock Deep anomaly detection on attributed networks.
\newblock In \emph{Proc. of SDM}.

\bibitem[{Duan et~al.(2022)Duan, Wang, Liu, Zhou, Hu, and
  Jin}]{jingcan2022gadmsl}
Duan, J.; Wang, S.; Liu, X.; Zhou, H.; Hu, J.; and Jin, H. 2022.
\newblock GADMSL: Graph Anomaly Detection on Attributed Networks via
  Multi-scale Substructure Learning.
\newblock \emph{arXiv preprint arXiv:2211.15255}.

\bibitem[{Feng et~al.(2020)Feng, Zhang, Dong, Han, Luan, Xu, Yang, Kharlamov,
  and Tang}]{feng2020graph}
Feng, W.; Zhang, J.; Dong, Y.; Han, Y.; Luan, H.; Xu, Q.; Yang, Q.; Kharlamov,
  E.; and Tang, J. 2020.
\newblock Graph random neural networks for semi-supervised learning on graphs.
\newblock \emph{Proc. of NeurIPS}.

\bibitem[{Garcia-Teodoro et~al.(2009)Garcia-Teodoro, Diaz-Verdejo,
  Maci{\'a}-Fern{\'a}ndez, and V{\'a}zquez}]{garcia2009anomaly}
Garcia-Teodoro, P.; Diaz-Verdejo, J.; Maci{\'a}-Fern{\'a}ndez, G.; and
  V{\'a}zquez, E. 2009.
\newblock Anomaly-based network intrusion detection: Techniques, systems and
  challenges.
\newblock \emph{computers \& security}.

\bibitem[{Hafidi et~al.(2022)Hafidi, Ghogho, Ciblat, and
  Swami}]{hafidi2022negative}
Hafidi, H.; Ghogho, M.; Ciblat, P.; and Swami, A. 2022.
\newblock Negative sampling strategies for contrastive self-supervised learning
  of graph representations.
\newblock \emph{Signal Processing}.

\bibitem[{Han et~al.(2022)Han, Hui, Jiang, Qian, and Xie}]{han2022generative}
Han, Y.; Hui, L.; Jiang, H.; Qian, J.; and Xie, J. 2022.
\newblock Generative Subgraph Contrast for Self-Supervised Graph Representation
  Learning.
\newblock In \emph{Proc. of ECCV}.

\bibitem[{Hassani and Khasahmadi(2020)}]{hassani2020contrastive}
Hassani, K.; and Khasahmadi, A.~H. 2020.
\newblock Contrastive multi-view representation learning on graphs.
\newblock In \emph{Proc. of ICML}.

\bibitem[{Hu et~al.(2022)Hu, Yu, Wang, Zhu, Cai, and Zhu}]{hu2022detecting}
Hu, J.; Yu, G.; Wang, S.; Zhu, E.; Cai, Z.; and Zhu, X. 2022.
\newblock Detecting Anomalous Events from Unlabeled Videos via Temporal Masked
  Auto-Encoding.
\newblock In \emph{Proc. of ICME}.

\bibitem[{Huang et~al.(2018)Huang, Mu, Yang, and Cai}]{huang2018codetect}
Huang, D.; Mu, D.; Yang, L.; and Cai, X. 2018.
\newblock CoDetect: Financial fraud detection with anomaly feature detection.
\newblock \emph{IEEE Access}.

\bibitem[{Huang et~al.(2021)Huang, Pei, Menkovski, and
  Pechenizkiy}]{huang2021hop}
Huang, T.; Pei, Y.; Menkovski, V.; and Pechenizkiy, M. 2021.
\newblock Hop-count based self-supervised anomaly detection on attributed
  networks.
\newblock \emph{arXiv preprint arXiv:2104.07917}.

\bibitem[{Jiao et~al.(2020)Jiao, Xiong, Zhang, Zhang, Zhang, and
  Zhu}]{jiao2020sub}
Jiao, Y.; Xiong, Y.; Zhang, J.; Zhang, Y.; Zhang, T.; and Zhu, Y. 2020.
\newblock Sub-graph contrast for scalable self-supervised graph representation
  learning.
\newblock In \emph{Proc. of ICDM}.

\bibitem[{Jin et~al.(2021{\natexlab{a}})Jin, Liu, Zheng, Chi, Li, and
  Pan}]{jin2021anemone}
Jin, M.; Liu, Y.; Zheng, Y.; Chi, L.; Li, Y.-F.; and Pan, S.
  2021{\natexlab{a}}.
\newblock ANEMONE: Graph Anomaly Detection with Multi-Scale Contrastive
  Learning.
\newblock In \emph{Proc. of CIKM}.

\bibitem[{Jin et~al.(2021{\natexlab{b}})Jin, Zheng, Li, Gong, Zhou, and
  Pan}]{jin2021multi}
Jin, M.; Zheng, Y.; Li, Y.-F.; Gong, C.; Zhou, C.; and Pan, S.
  2021{\natexlab{b}}.
\newblock Multi-scale contrastive siamese networks for self-supervised graph
  representation learning.
\newblock In \emph{Proc. of IJCAI}.

\bibitem[{Kipf and Welling(2016)}]{kipf2016semi}
Kipf, T.~N.; and Welling, M. 2016.
\newblock Semi-supervised classification with graph convolutional networks.
\newblock \emph{arXiv preprint arXiv:1609.02907}.

\bibitem[{Klicpera, Wei{\ss}enberger, and
  G{\"u}nnemann(2019)}]{klicpera2019diffusion}
Klicpera, J.; Wei{\ss}enberger, S.; and G{\"u}nnemann, S. 2019.
\newblock Diffusion improves graph learning.
\newblock In \emph{Proc. of ICONIP}.

\bibitem[{Li et~al.(2017)Li, Dani, Hu, and Liu}]{li2017radar}
Li, J.; Dani, H.; Hu, X.; and Liu, H. 2017.
\newblock Radar: Residual Analysis for Anomaly Detection in Attributed
  Networks.
\newblock In \emph{Proc. of IJCAI}.

\bibitem[{Liang et~al.(2022)Liang, Liu, Zhou, Liu, and Tu}]{KGESymCL}
Liang, K.; Liu, Y.; Zhou, S.; Liu, X.; and Tu, W. 2022.
\newblock Relational Symmetry based Knowledge Graph Contrastive Learning.

\bibitem[{Liu et~al.(2022{\natexlab{a}})Liu, Jin, Pan, Zhou, Zheng, Xia, and
  Yu}]{liu2022graph}
Liu, Y.; Jin, M.; Pan, S.; Zhou, C.; Zheng, Y.; Xia, F.; and Yu, P.
  2022{\natexlab{a}}.
\newblock Graph self-supervised learning: A survey.
\newblock \emph{IEEE Transactions on Knowledge and Data Engineering}.

\bibitem[{Liu et~al.(2021)Liu, Li, Pan, Gong, Zhou, and
  Karypis}]{liu2021anomaly}
Liu, Y.; Li, Z.; Pan, S.; Gong, C.; Zhou, C.; and Karypis, G. 2021.
\newblock Anomaly detection on attributed networks via contrastive
  self-supervised learning.
\newblock \emph{IEEE transactions on neural networks and learning systems}.

\bibitem[{Liu et~al.(2022{\natexlab{b}})Liu, Tu, Zhou, Liu, Song, Yang, and
  Zhu}]{liu2022deep}
Liu, Y.; Tu, W.; Zhou, S.; Liu, X.; Song, L.; Yang, X.; and Zhu, E.
  2022{\natexlab{b}}.
\newblock Deep Graph Clustering via Dual Correlation Reduction.
\newblock In \emph{Proc. of AAAI}.

\bibitem[{Liu et~al.(2022{\natexlab{c}})Liu, Xia, Zhou, Wang, Guo, Yang, Liang,
  Tu, Li, and Liu}]{yue2022survey}
Liu, Y.; Xia, J.; Zhou, S.; Wang, S.; Guo, X.; Yang, X.; Liang, K.; Tu, W.; Li,
  Z.~S.; and Liu, X. 2022{\natexlab{c}}.
\newblock A Survey of Deep Graph Clustering: Taxonomy, Challenge, and
  Application.
\newblock \emph{arXiv preprint arXiv:2211.12875}.

\bibitem[{Liu et~al.(2022{\natexlab{d}})Liu, Yang, Zhou, and
  Liu}]{liu2022simple}
Liu, Y.; Yang, X.; Zhou, S.; and Liu, X. 2022{\natexlab{d}}.
\newblock Simple Contrastive Graph Clustering.
\newblock \emph{arXiv preprint arXiv:2205.07865}.

\bibitem[{Liu et~al.(2022{\natexlab{e}})Liu, Zhou, Liu, Tu, and
  Yang}]{liu2022improved}
Liu, Y.; Zhou, S.; Liu, X.; Tu, W.; and Yang, X. 2022{\natexlab{e}}.
\newblock Improved Dual Correlation Reduction Network.
\newblock \emph{arXiv preprint arXiv:2202.12533}.

\bibitem[{Ma et~al.(2021)Ma, Wu, Xue, Yang, Zhou, Sheng, Xiong, and
  Akoglu}]{ma2021comprehensive}
Ma, X.; Wu, J.; Xue, S.; Yang, J.; Zhou, C.; Sheng, Q.~Z.; Xiong, H.; and
  Akoglu, L. 2021.
\newblock A comprehensive survey on graph anomaly detection with deep learning.
\newblock \emph{IEEE Transactions on Knowledge and Data Engineering}.

\bibitem[{Mrabah et~al.(2022)Mrabah, Bouguessa, Touati, and
  Ksantini}]{mrabah2022rethinking}
Mrabah, N.; Bouguessa, M.; Touati, M.~F.; and Ksantini, R. 2022.
\newblock Rethinking graph auto-encoder models for attributed graph clustering.
\newblock \emph{IEEE Transactions on Knowledge and Data Engineering}.

\bibitem[{M{\"u}ller et~al.(2013)M{\"u}ller, S{\'a}nchez, M{\"u}lle, and
  B{\"o}hm}]{muller2013ranking}
M{\"u}ller, E.; S{\'a}nchez, P.~I.; M{\"u}lle, Y.; and B{\"o}hm, K. 2013.
\newblock Ranking outlier nodes in subspaces of attributed graphs.
\newblock In \emph{2013 IEEE 29th international conference on data engineering
  workshops (ICDEW)}.

\bibitem[{Oord, Li, and Vinyals(2018)}]{oord2018representation}
Oord, A. v.~d.; Li, Y.; and Vinyals, O. 2018.
\newblock Representation learning with contrastive predictive coding.
\newblock \emph{arXiv preprint arXiv:1807.03748}.

\bibitem[{Peng et~al.(2018)Peng, Luo, Li, Liu, and Zheng}]{peng2018anomalous}
Peng, Z.; Luo, M.; Li, J.; Liu, H.; and Zheng, Q. 2018.
\newblock ANOMALOUS: A Joint Modeling Approach for Anomaly Detection on
  Attributed Networks.
\newblock In \emph{Proc. of IJCAI}.

\bibitem[{Perozzi and Akoglu(2016)}]{perozzi2016scalable}
Perozzi, B.; and Akoglu, L. 2016.
\newblock Scalable anomaly ranking of attributed neighborhoods.
\newblock In \emph{Proc. of SDM}.

\bibitem[{Perozzi et~al.(2014)Perozzi, Akoglu, Iglesias~S{\'a}nchez, and
  M{\"u}ller}]{perozzi2014focused}
Perozzi, B.; Akoglu, L.; Iglesias~S{\'a}nchez, P.; and M{\"u}ller, E. 2014.
\newblock Focused clustering and outlier detection in large attributed graphs.
\newblock In \emph{Proc. of KDD}.

\bibitem[{Qiu et~al.(2020)Qiu, Chen, Dong, Zhang, Yang, Ding, Wang, and
  Tang}]{qiu2020gcc}
Qiu, J.; Chen, Q.; Dong, Y.; Zhang, J.; Yang, H.; Ding, M.; Wang, K.; and Tang,
  J. 2020.
\newblock Gcc: Graph contrastive coding for graph neural network pre-training.
\newblock In \emph{Proc. of KDD}.

\bibitem[{Sen et~al.(2008)Sen, Namata, Bilgic, Getoor, Galligher, and
  Eliassi-Rad}]{sen2008collective}
Sen, P.; Namata, G.; Bilgic, M.; Getoor, L.; Galligher, B.; and Eliassi-Rad, T.
  2008.
\newblock Collective classification in network data.
\newblock \emph{AI magazine}.

\bibitem[{Tang et~al.(2022)Tang, Li, Gao, and Li}]{tang2022rethinking}
Tang, J.; Li, J.; Gao, Z.; and Li, J. 2022.
\newblock Rethinking Graph Neural Networks for Anomaly Detection.
\newblock \emph{arXiv preprint arXiv:2205.15508}.

\bibitem[{Thakoor et~al.(2021)Thakoor, Tallec, Azar, Munos,
  Veli{\v{c}}kovi{\'c}, and Valko}]{thakoor2021bootstrapped}
Thakoor, S.; Tallec, C.; Azar, M.~G.; Munos, R.; Veli{\v{c}}kovi{\'c}, P.; and
  Valko, M. 2021.
\newblock Bootstrapped Representation Learning on Graphs.
\newblock In \emph{ICLR 2021 Workshop on Geometrical and Topological
  Representation Learning}.

\bibitem[{Tu et~al.(2021)Tu, Zhou, Liu, Guo, Cai, zhu, and Cheng}]{DFCN2021}
Tu, W.; Zhou, S.; Liu, X.; Guo, X.; Cai, Z.; zhu, E.; and Cheng, J. 2021.
\newblock Deep Fusion Clustering Network.
\newblock In \emph{Proc. of AAAI}.

\bibitem[{Tu et~al.(2022)Tu, Zhou, Liu, Yue, Cai, zhu, Changwang, and
  Cheng}]{ITR2022}
Tu, W.; Zhou, S.; Liu, X.; Yue, L.; Cai, Z.; zhu, E.; Changwang, Z.; and Cheng,
  J. 2022.
\newblock Initializing Then Refining: A Simple Graph Attribute Imputation
  Network.
\newblock In \emph{Proc. of IJCAI}.

\bibitem[{Velickovic et~al.(2019)Velickovic, Fedus, Hamilton, Li{\`o}, Bengio,
  and Hjelm}]{velickovic2019deep}
Velickovic, P.; Fedus, W.; Hamilton, W.~L.; Li{\`o}, P.; Bengio, Y.; and Hjelm,
  R.~D. 2019.
\newblock Deep Graph Infomax.
\newblock \emph{ICLR (Poster)}.

\bibitem[{Wang et~al.(2018)Wang, Liu, Jiao, Chen, and Jin}]{wang2018unified}
Wang, W.; Liu, X.; Jiao, P.; Chen, X.; and Jin, D. 2018.
\newblock A unified weakly supervised framework for community detection and
  semantic matching.
\newblock In \emph{Proc. of KDD}.

\bibitem[{Wang et~al.(2021)Wang, Wang, Liang, Cai, and Hooi}]{wang2021mixup}
Wang, Y.; Wang, W.; Liang, Y.; Cai, Y.; and Hooi, B. 2021.
\newblock Mixup for node and graph classification.
\newblock In \emph{Proc. of WWW}.

\bibitem[{Wu et~al.(2019)Wu, Morstatter, Carley, and
  Liu}]{wu2019misinformation}
Wu, L.; Morstatter, F.; Carley, K.~M.; and Liu, H. 2019.
\newblock Misinformation in social media: definition, manipulation, and
  detection.
\newblock \emph{ACM SIGKDD Explorations Newsletter}.

\bibitem[{Wu et~al.(2020)Wu, Pan, Chen, Long, Zhang, and
  Philip}]{wu2020comprehensive}
Wu, Z.; Pan, S.; Chen, F.; Long, G.; Zhang, C.; and Philip, S.~Y. 2020.
\newblock A comprehensive survey on graph neural networks.
\newblock \emph{IEEE transactions on neural networks and learning systems}.

\bibitem[{Xu et~al.(2007)Xu, Yuruk, Feng, and Schweiger}]{xu2007scan}
Xu, X.; Yuruk, N.; Feng, Z.; and Schweiger, T.~A. 2007.
\newblock Scan: a structural clustering algorithm for networks.
\newblock In \emph{Proc. of KDD}.

\bibitem[{Yang et~al.(2022{\natexlab{a}})Yang, Hu, Zhou, Liu, and
  Zhu}]{9817089}
Yang, X.; Hu, X.; Zhou, S.; Liu, X.; and Zhu, E. 2022{\natexlab{a}}.
\newblock Interpolation-Based Contrastive Learning for Few-Label
  Semi-Supervised Learning.
\newblock \emph{IEEE Transactions on Neural Networks and Learning Systems}.

\bibitem[{Yang et~al.(2022{\natexlab{b}})Yang, Liu, Zhou, Liu, and
  Zhu}]{yang2022interpolation}
Yang, X.; Liu, Y.; Zhou, S.; Liu, X.; and Zhu, E. 2022{\natexlab{b}}.
\newblock Mixed Graph Contrastive Network for Semi-Supervised Node
  Classification.
\newblock \emph{arXiv preprint arXiv:2206.02796}.

\bibitem[{You et~al.(2020)You, Chen, Sui, Chen, Wang, and Shen}]{you2020graph}
You, Y.; Chen, T.; Sui, Y.; Chen, T.; Wang, Z.; and Shen, Y. 2020.
\newblock Graph contrastive learning with augmentations.
\newblock \emph{Proc. of NeurIPS}.

\bibitem[{Yuan et~al.(2021)Yuan, Zhou, Yu, Huang, Chen, and
  Xia}]{yuan2021higher}
Yuan, X.; Zhou, N.; Yu, S.; Huang, H.; Chen, Z.; and Xia, F. 2021.
\newblock Higher-order Structure Based Anomaly Detection on Attributed
  Networks.
\newblock In \emph{2021 IEEE International Conference on Big Data (Big Data)}.

\bibitem[{Zbontar et~al.(2021)Zbontar, Jing, Misra, LeCun, and
  Deny}]{zbontar2021barlow}
Zbontar, J.; Jing, L.; Misra, I.; LeCun, Y.; and Deny, S. 2021.
\newblock Barlow twins: Self-supervised learning via redundancy reduction.
\newblock In \emph{Proc. of ICML}.

\bibitem[{Zhang, Wang, and Chen(2022)}]{zhang2022reconstruction}
Zhang, J.; Wang, S.; and Chen, S. 2022.
\newblock Reconstruction Enhanced Multi-View Contrastive Learning for Anomaly
  Detection on Attributed Networks.
\newblock \emph{arXiv preprint arXiv:2205.04816}.

\bibitem[{Zhao et~al.(2022)Zhao, Liu, G{\"u}nnemann, and Jiang}]{zhao2022graph}
Zhao, T.; Liu, G.; G{\"u}nnemann, S.; and Jiang, M. 2022.
\newblock Graph Data Augmentation for Graph Machine Learning: A Survey.
\newblock \emph{arXiv preprint arXiv:2202.08871}.

\bibitem[{Zhao et~al.(2021)Zhao, Liu, Neves, Woodford, Jiang, and
  Shah}]{zhao2021data}
Zhao, T.; Liu, Y.; Neves, L.; Woodford, O.; Jiang, M.; and Shah, N. 2021.
\newblock Data augmentation for graph neural networks.
\newblock In \emph{Proc. of AAAI}.

\bibitem[{Zheng et~al.(2021)Zheng, Jin, Liu, Chi, Phan, and
  Chen}]{zheng2021generative}
Zheng, Y.; Jin, M.; Liu, Y.; Chi, L.; Phan, K.~T.; and Chen, Y.-P.~P. 2021.
\newblock Generative and Contrastive Self-Supervised Learning for Graph Anomaly
  Detection.
\newblock \emph{IEEE Transactions on Knowledge and Data Engineering}.

\bibitem[{Zhou et~al.(2021)Zhou, Tan, Xu, Huang, and
  Chung}]{zhou2021subtractive}
Zhou, S.; Tan, Q.; Xu, Z.; Huang, X.; and Chung, F.-l. 2021.
\newblock Subtractive Aggregation for Attributed Network Anomaly Detection.
\newblock In \emph{Proc. of CIKM}.

\bibitem[{Zhu et~al.(2022)Zhu, Guo, Wu, and Tang}]{zhu2022rosa}
Zhu, Y.; Guo, J.; Wu, F.; and Tang, S. 2022.
\newblock RoSA: A Robust Self-Aligned Framework for Node-Node Graph Contrastive
  Learning.
\newblock \emph{arXiv preprint arXiv:2204.13846}.

\end{thebibliography}

\end{document}